\documentclass[runningheads]{llncs}

\usepackage{graphicx}
\usepackage{amsmath}
\usepackage{amssymb}
\usepackage{nicefrac}

\usepackage{mathtools}
\usepackage{ntheorem}

\renewcommand{\qed}{\hfill{$\square$}}

\newtheorem*{definition*}{Definition}

\usepackage{tabularx}
\usepackage{booktabs}

\newcolumntype{P}[1]{>{\centering\arraybackslash}p{#1}}
\newcolumntype{R}{>{\raggedleft\arraybackslash}X}
\newcolumntype{C}{>{\centering\arraybackslash}X}

\newcommand{\dv}[2]{\frac{d#1}{d#2}}
\newcommand{\dvtext}[2]{\nicefrac{d#1}{d#2}}
\newcommand{\cdv}[2]{\frac{\partial#1}{\partial#2}}
\newcommand{\cdvtext}[2]{\nicefrac{\partial#1}{\partial#2}}
\newcommand{\loss}[0]{\mathcal{L}}
\newcommand{\ovar}[0]{h}
\newcommand{\hvar}[0]{c}

\newcommand{\oimvar}[0]{\beta}
\newcommand{\oeltra}[0]{b}

\DeclareMathSizes{10}{9}{7}{5}

\begin{document}

\title{A Taxonomy of Recurrent Learning Rules}

\author{Guillermo Martín-Sánchez\inst{1} \and
Sander Bohté\inst{3} \and
Sebastian Otte\inst{1,2}}
\authorrunning{G. Martín-Sánchez et al.}

\institute{Neuro-Cognitive Modeling, University of Tübingen\\Sand 14, 72076 Tübingen, Germany\\
\email{guillemartinsan@gmail.com}\\ \and
Adaptive AI Lab, University of Lübeck\\Ratzeburger Allee 160, 23562 Lübeck, Germany\\
\email{sebastian.otte@uni-luebeck.de}\\ \and
Machine Learning group, CWI\\
Science Park 123, NL-1098XG Amsterdam, The Netherlands\\
\email{S.M.Bohte@cwi.nl}
}
\maketitle         
\begin{abstract} Backpropagation through time (BPTT) is the de facto standard for training recurrent neural networks (RNNs), but
it is non-causal and non-local. Real-time recurrent learning is a causal alternative, but it is highly inefficient. Recently, e-prop was proposed as a causal, local, and efficient practical alternative to these algorithms, providing an approximation of the exact gradient by radically pruning the recurrent dependencies carried over time.
Here, we derive RTRL from BPTT using a detailed notation bringing intuition and clarification to how they are connected. Furthermore, we frame e-prop within in the picture, formalising what it approximates. Finally, we derive a family of algorithms of which e-prop is a special case.
\end{abstract}

\addtolength{\textfloatsep}{-0.3cm}
\addtolength{\abovecaptionskip}{-0.25cm}
\addtolength{\belowcaptionskip}{-0.25cm}

\newcommand{\highlight}[1]{\textbf{#1}}

\section{Introduction}

\emph{Backpropagation through time} ({BPTT}) \cite{Werbos1990} is currently the most used algorithm for training \emph{recurrent neural networks} (RNNs) and is derived from applying the chain rule (backpropagation) to the computational graph of the RNN unrolled in time. It suffers however from undesired characteristics both in terms of biological plausibility and large scale applicability: (i) it is \emph{non-causal}, since at each time step it requires future activity to compute the current gradient of the loss with respect to the parameters; and (ii) it is \emph{non-local}, since it requires reverse error signal propagating across all neurons and all synapses. 
An equivalent algorithm is \emph{real-time recurrent learning} ({RTRL}) \cite{Williams1989}. It uses eligibility traces that are computed at each time step recursively in order to be causal, and can therefore be computed online. However, this comes at the cost of very high computational and memory complexity, since all temporal forward dependencies have to be maintained over time. RTRL is, hence, also non-local.
Recently, a new online learning algorithm, called \emph{e-prop} \cite{Bellec2020} has been proposed, which is tailored for training \emph{recurrent spiking neural networks} (RSNNs) with local neural dynamics. The aim was to find an alternative to BPTT (and RTRL)  that is causal, local, but also computational and memory efficient. 

\begin{figure}[t!]
\centering
\includegraphics[width=0.8\textwidth]{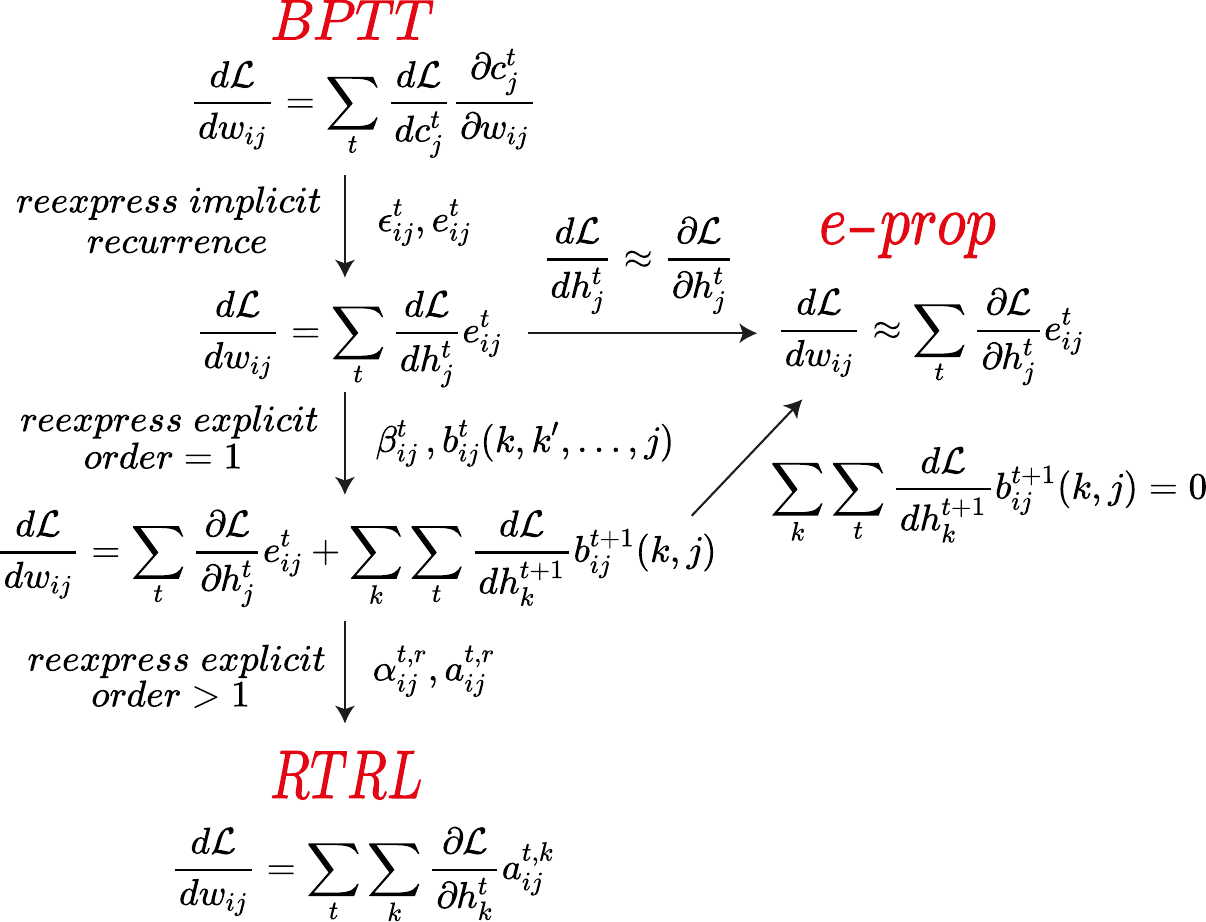} 
\caption{Overview of all the algorithms and how they relate to each other.}
\label{fig:overview}
\end{figure}

In this paper, we look in depth into the formalisation of BPTT and RTRL and formalise e-prop into the picture. To do so, we use the computational graph and notation of the architecture in the e-prop paper \cite{Bellec2020} to understand how these three algorithms relate to each other. Furthermore, in a posterior paper \cite{Zenke2020}, it was shown that e-prop was an approximation of RTRL. Here, by formalising also RTRL in the same framework we indeed confirm the connection and make it more explicit (cf. Fig.~\ref{fig:overview}). In the process, we uncover a family of algorithms determined by the level of approximation allowed to benefit from causality and locality. The main focus of this paper is to give intuition and understanding of all of these gradient computation rules.

\subsection{Background}

The most common way to train a model in supervised learning is to compute the gradient of a given loss $\loss$ with respect to the parameters $\theta$, $\dvtext{\loss}{\theta}$, and use this gradient in some gradient descent scheme of the form $\theta(\tau+1) = \theta(\tau) - f(\dvtext{\loss}{\theta})$, where $\tau$ refers to the current update iteration and $f$ is some gradient postprocessing. Therefore, we here focus on the algorithms for the computation (or approximation) of this gradient.

In particular, we focus on a general class of RNN models where we have $n$ computational units. These units have hidden states at each time step $\hvar_i^t$ that influence the hidden state at the next time step $\hvar_i^{t+1}$ (implicit recurrence) as well as the output of the unit at the current time step $\ovar_i^t$. The output $\ovar_i^t$ of a unit at a given time step influences the hidden state of the same and other units at the following time step $\hvar_j^{t+1}$ (explicit recurrence) through a weighted connection $w_{ij}$. Finally, these outputs also account for the model's computation (either directly or through some other computations, e.g. a linear readout) and therefore are subject to evaluation by a loss function $\loss$. The formalization here is agnostic to the particular dimensionality and computational relation between the variables and therefore apply for different RNNs, such as LSTMs \cite{hochreiter1997long} or RSNNs \cite{Bellec2018}.

\begin{figure}[t!]
\centering
\includegraphics[scale=0.45]{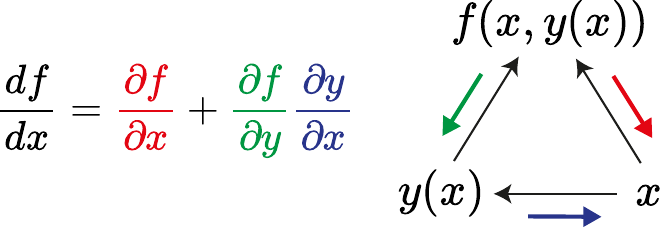} 
\caption{Simple example of computational graph and distinction between total and partial derivative of $f$ with respect to $x$.} \label{fig:notation}
\end{figure} 

For a function $f(x,y(x))$ we distinguish the notation of the total derivative $\dvtext{f}{x}$ and the partial derivative $\cdvtext{f}{x}$ because the first one represents the whole gradient through all paths, while the second one expresses only the direct relation between the variables. To illustrate: using the chain rule (cf. Fig.~\ref{fig:notation}) and with the example $y = 2x$ and $f(x,y(x)) = xy$, the total derivative is calculated as:
\begin{equation}
    \dv{f}{x} = \cdv{f}{x} + \cdv{f}{y}\cdv{y}{x} = y + x\cdot 2 = 4x
\end{equation}

\section{Backpropagation Through Time}

In RNNs, since previous states affect the current state, the trick to applying the chain rule is to unroll the RNN in time, obtain a virtual feed-forward architecture and apply to this computational graph error-backpropagation \cite{Werbos1974}. The resulting algorithm is BPTT \cite{Werbos1990} and it is the currently most used algorithm to compute \smash{$\dvtext{\loss}{w_{ij}}$} since it reuses many previous computations to be highly efficient. Here, we focus our attention on the role of the recurrences dividing the algorithm into the following steps:

\begin{figure}[t!]
\includegraphics[width=\textwidth]{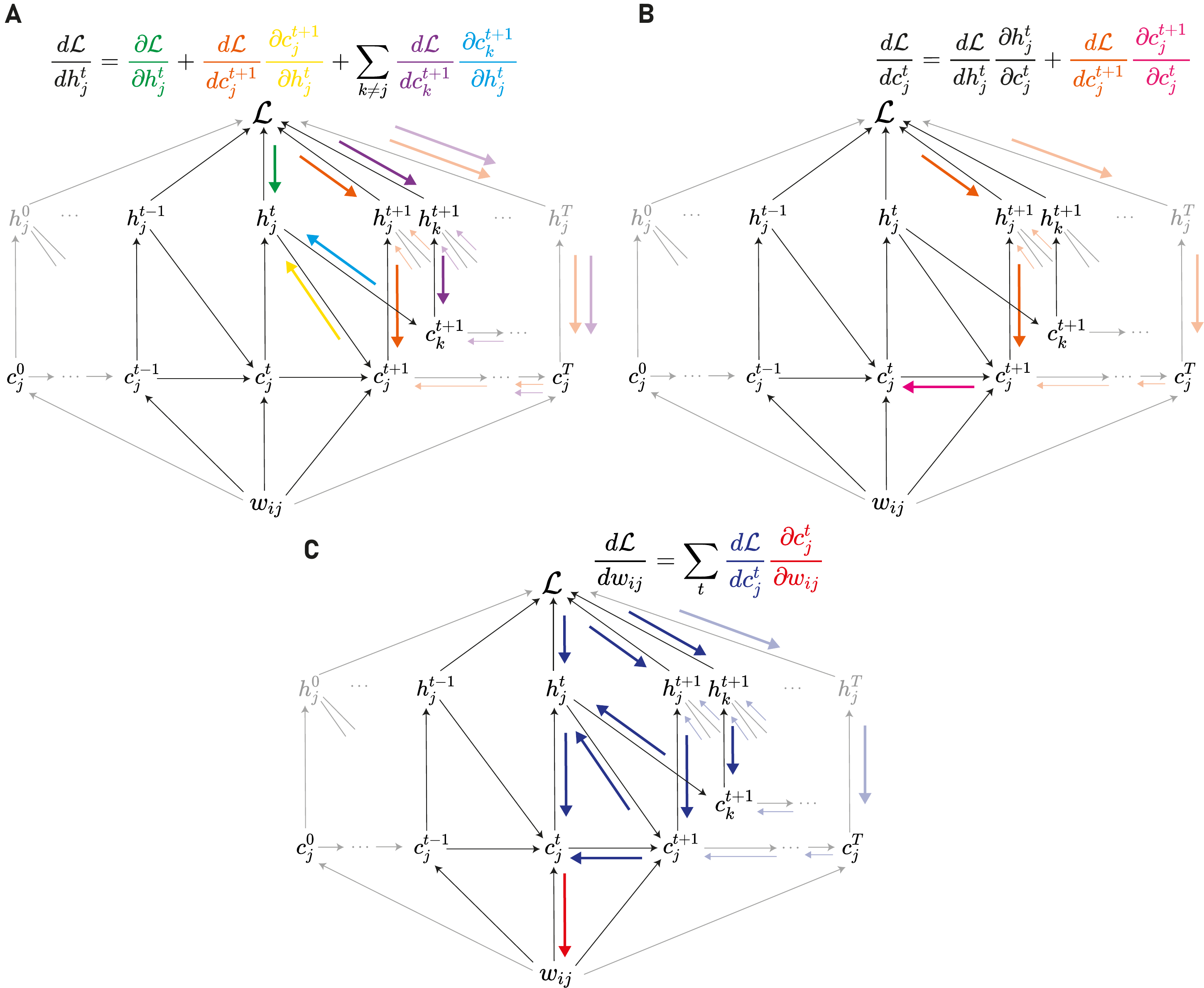} 
\caption{Computational graph for A) explicit recurrences gradients, B) implicit recurrence gradients and C) final computation of BPTT.} \label{fig:EandIandB}
\end{figure}

\subsubsection{Explicit Recurrences:} Compute \smash{$\dvtext{\loss}{\ovar^{t}_j}$} using the recursive definition given by the explicit recurrences (cf. Fig.~\ref{fig:EandIandB}A):
    \begin{equation}\label{eqdedz}
        \begin{split}
    \dv{\loss}{\ovar^{t}_j} &= \cdv{ \loss}{ \ovar^{t}_j} +  \dv{\loss}{\hvar^{t+1}_j} \cdv{ \hvar^{t+1}_j}{ \ovar_j^{t}} + \sum_{k \neq j} \dv{\loss}{\hvar^{t+1}_k} \cdv{ \hvar^{t+1}_k}{ \ovar_j^{t}} \\
    &= \cdv{ \loss}{ \ovar^{t}_j} + \sum_k \dv{\loss}{\hvar^{t+1}_k} \cdv{ \hvar^{t+1}_k}{ \ovar_j^{t}}
    \end{split}
    \end{equation}
    
\subsubsection{Implicit Recurrences:} Compute \smash{$\dvtext{\loss}{\hvar^{t}_j}$} using the value of the previous step and the recursive definition given by the implicit recurrence (cf. Fig.~\ref{fig:EandIandB}B):
    \begin{equation}\label{eqdedh}
    \dv{\loss}{\hvar^{t}_j} = \dv{\loss}{\ovar^{t}_j} \cdv{ \ovar^{t}_j}{ \hvar_j^{t}} + \dv{\loss}{\hvar^{t+1}_j}\cdv{ \hvar_j^{t+1}}{ \hvar_j^{t}}
    \end{equation}
    
\subsubsection{BPTT:} Finally, compute \smash{$\dvtext{\loss}{w_{ij}}$} using the values  obtained in the previous two steps for all time steps (cf. Fig.~\ref{fig:EandIandB}C):
    \begin{equation}\label{eqg1}
    \dv{\loss}{w_{ij}} = \sum_{t} \dv{\loss}{\hvar_j^{t}} \cdv{ \hvar^{t}_j}{ w_{ij}}
    \end{equation}

We use explicit and implicit recurrences from the maximum time $T$ backwards and for all $t \leq T$, and finally, sum all the results from Eq.~\ref{eqdedh}. The existence of these recurrences makes BPTT present the following problems \cite{Marblestone2016,Bellec2020}:

\subsubsection{Non-locality:}
Due to the explicit recurrences, we need to take into account how the current synaptic strength $w_{ij}$ between the neurons $i$ and $j$ affects the future value of the postsynaptic neuron: \smash{$\cdvtext{ \hvar^{t+1}_k}{ \ovar^t_j}$} for all $k \neq j$ (cf. Eq.~\ref{eqdedz}). This means that to compute the weight change for synapse magnitude $w_{ij}$ we need information of the hidden variables $\hvar^{t+1}_k$ for all $k$. Moreover, this chain of dependencies continues at each time step, such that at the next time step we need information of the variables $\hvar^{t+2}_q$ for all $q$ (including $q \neq j$) and so forth. The contraposition would be a \textbf{local} algorithm that does not require messages passing from every neuron to every synapse to compute the gradients, but rather only need information close to the given synapse.

\subsubsection{Non-causality:}
Due to the three kinds of recurrences shown before we need to take into account all the gradients in the future (the same way as current layer computations need to use the gradients of posterior layers in feed-forward architectures), leading to two main problems. First, we need to compute the values of the variables (update locking) and the gradients (backwards locking) across all future time steps before computing the current gradient \cite{Czarnecki2017}. Secondly, all the values of all the variables across time have to be saved during the inference phase to be used while computing the gradients, requiring a memory overhead ($O(nT)$ with $n$ neurons and $T$ time steps). The contraposition would be a \textbf{causal} algorithm, that at each time step would only need information from previous and current activity to compute the current gradient. Therefore, it could do it at each time step (\textbf{online}) and while the inference is running.

\section{Real-Time Recurrent Learning}

RTRL \cite{Williams1989} is a causal learning algorithm that can be implemented as an online learning algorithm and that computes the same gradient as BPTT, at the cost of being more computationally expensive. 
We derive the equation for RTRL starting with BPTT (cf. Eq.~\ref{eqg1}) via re-expressing the gradients that connect the computation with future gradients to obtain a causal algorithm. These gradients correspond to the implicit and explicit recurrences.

\subsection{Re-expressing Implicit Recurrence}
\label{sect:ReexImplicitRec}

%\subsubsection{Considering the implicit recurrence.}
First, we re-express the implicit recurrence gradient \smash{$\cdvtext{ \hvar_j^{t+1}}{ \hvar_j^{t}}$}. 

\subsubsection{Unrolling the recursion:} To unroll, we plug the equation of implicit recurrence Eq.~\ref{eqdedh} into Eq.~\ref{eqg1}:
\begin{equation}\label{eqimroll}
\begin{split}
    \dv{\loss}{w_{ij}} =& \sum_{t'} \left( \dv{\loss}{\ovar^{t'}_j} \cdv{ \ovar^{t'}_j}{ \hvar_j^{t'}} + \dv{\loss}{\hvar^{t'+1}_j}\cdv{ \hvar_j^{t'+1}}{ \hvar_j^{t'}} \right) \cdv{ \hvar^{t'}_j}{ w_{ij}} \\
    =&\sum_{t'} \left( \dv{\loss}{\ovar^{t'}_j} \cdv{ \ovar^{t'}_j}{ \hvar_j^{t'}} + \left( \dv{\loss}{\ovar^{t'+1}_j} \cdv{ \ovar^{t'+1}_j}{ \hvar_j^{t'+1}} + (\cdots) \cdv{ \hvar_j^{t'+2}}{ \hvar_j^{t'+1}} \right) \cdv{ \hvar_j^{t'+1}}{ \hvar_j^{t'}} \right) \cdv{ \hvar^{t'}_j}{ w_{ij}}  \\
    =&\sum_{t'}\sum_{t \geq t'} \dv{\loss}{\ovar^t_j} \cdv{ \ovar^t_j}{ \hvar^t_j} \cdv{ \hvar^t_j}{ \hvar^{t-1}_j} \cdots \cdv{ \hvar^{t'+1}_j}{ \hvar^{t'}_j} \cdv{ \hvar^{t'}_j}{ w_{ij}}
\end{split}
\end{equation}

\subsubsection{Flip time indices:} The derived formula is non-causal since it requires future gradients (for each $t'$ we sum products of gradients with factors starting from $t \geq t'$). To make it causal, we change the indices as follows:
\begin{equation}\label{eqichang}
\begin{split}
    \dv{\loss}{w_{ij}} =&\sum_t \dv{\loss}{\ovar^t_j} \cdv{ \ovar^t_j}{ \hvar^t_j} \sum_{t' \leq t} \cdv{ \hvar^t_j}{ \hvar^{t-1}_j} \cdots \cdv{ \hvar^{t'+1}_j}{ \hvar^{t'}_j}  \cdv{ \hvar^{t'}_j}{ w_{ij}}
\end{split}
\end{equation}

\begin{figure}[t!]
\centering
\includegraphics[width=0.75\linewidth]{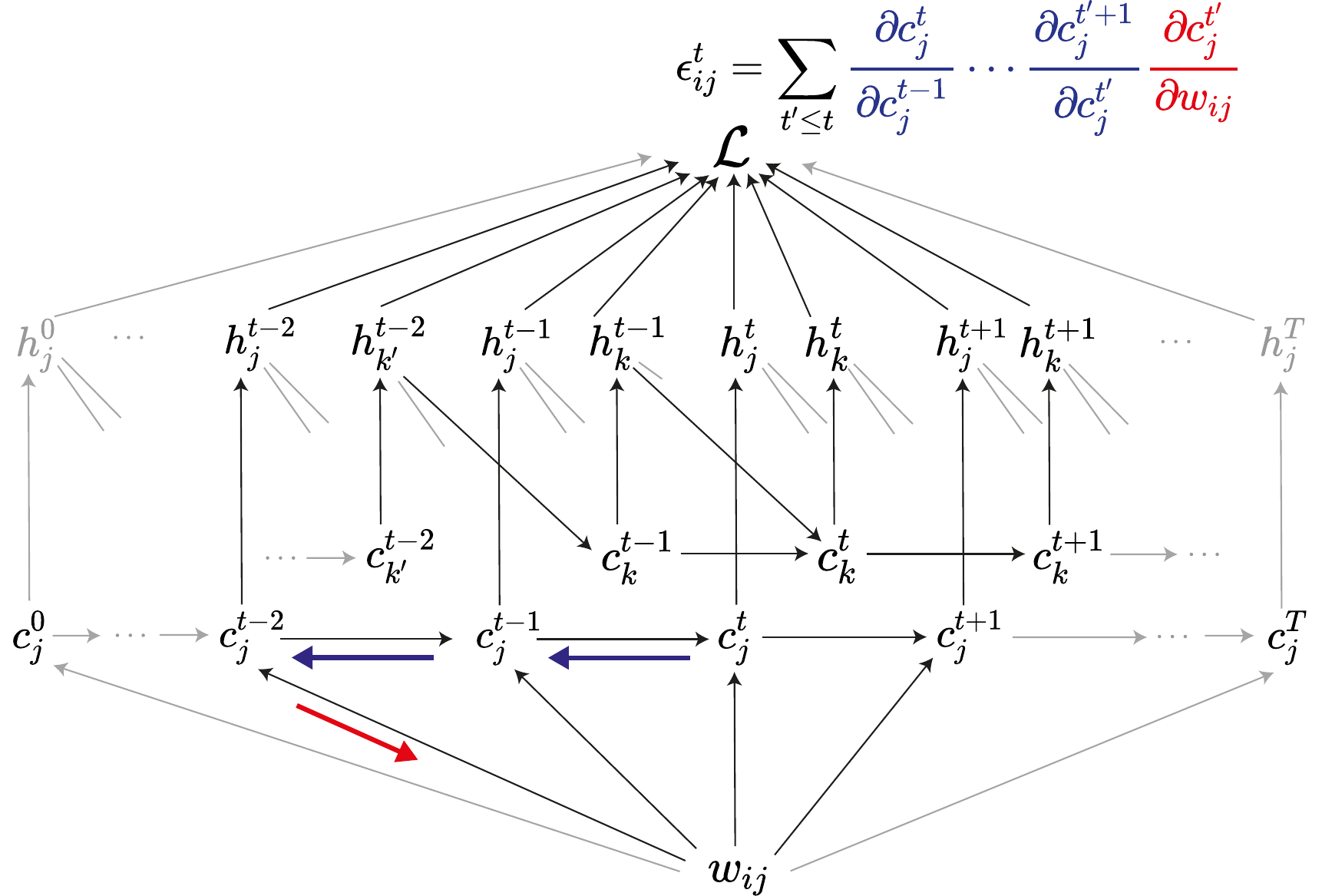} 
\caption{Computational graph for the implicit variable $\epsilon^t_{ij}$ with $t'= t-2$.}
\label{fig:IET}
\end{figure}

\begin{definition*}[Implicit variable]
We define the \emph{implicit variable} $\epsilon^t_{ij}$ as:
\begin{equation}\label{eqiet}
     \epsilon_{ij}^t := \sum_{t' \leq t} \cdv{ \hvar^t_j}{ \hvar^{t-1}_j} \cdots \cdv{ \hvar^{t'+1}_j}{ \hvar^{t'}_j} \cdv{ \hvar^{t'}_j}{ w_{ij}}
\end{equation}

\end{definition*}

\paragraph{Backwards interpretation:} Starting at $\hvar^t_j$, the implicit variable represents the sum over all the paths going backwards through the implicit recurrence until $\hvar^{t'}_j$ and from there to the synaptic weight $w_{ij}$ (cf. Fig.~\ref{fig:IET}).

\paragraph{Forwards interpretation:} The implicit variable represents how the hidden variable of neuron $j$ has been affected by the synapse weight $w_{ij}$ through time, i.e. taking into account also how the hidden variables at previous time steps have affected the variables at the current time step through the implicit recurrence. 

\paragraph{Incremental computation:} Importantly, there is a recursive relation to this variable that allows it to be updated at each time step:
\begin{equation}\label{eqincremiet}
     \epsilon_{ij}^t = \cdv{ \hvar^t_j}{ \hvar^{t-1}_j}  \epsilon^{t-1}_{ij} + \cdv{ \hvar^t_j}{ w_{ij}}
\end{equation}

\begin{definition*}[Implicit eligibility trace]
%\paragraph{Output trace:} 
Given the implicit variable $\epsilon^t_{ij}$, we define the \emph{implicit eligibility trace} $e^t_{ij}$ as:
\begin{equation}
    e^t_{ij} := \cdv{ \ovar^{t}_j}{ \hvar^{t}_j} \epsilon^t_{ij}
\end{equation} 
\end{definition*}

Since \smash{$\cdvtext{ \ovar^{t}_j}{ \hvar^{t}_j}$} is causal and local, and so is the implicit variable $\epsilon^t_{ij}$ (can be computed at each time step and is specific for each synapse), then the implicit eligibility trace $e^t_{ij}$ is also \textbf{causal} and \textbf{local}.

\subsubsection{Final equation with re-expressed implicit recurrence:}
\begin{figure}[t!]
\includegraphics[width=0.95\textwidth]{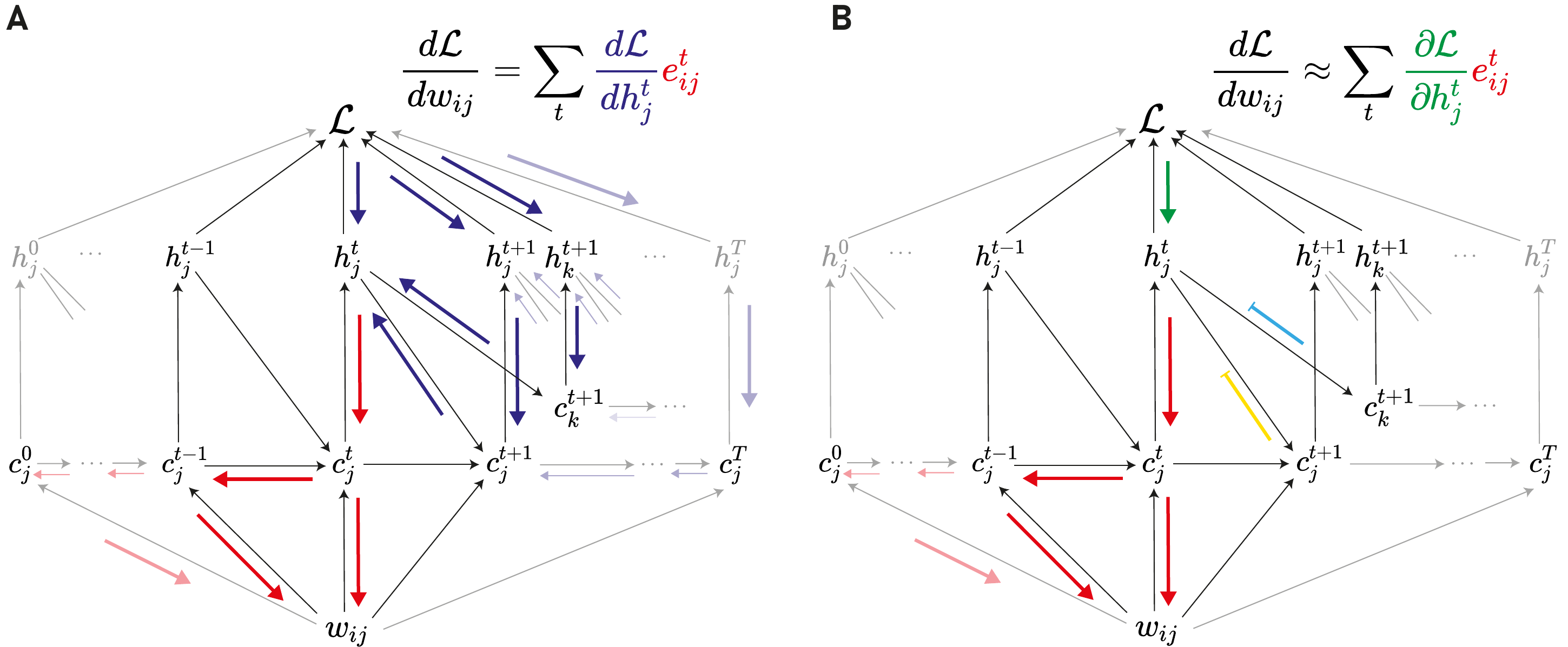} 
\caption{Computational graph for A) BPTT re-expressed with implicit elegibility trace (cf. Eq.~\ref{eqg2}) B) symmetric e-prop.}
\label{fig:eprop}
\end{figure}

With all of this combined, BPTT (cf. Eq.~\ref{eqg1}) has become (substituting Eq.~\ref{eqiet} in Eq.~\ref{eqichang}) the following (cf. Fig.~\ref{fig:eprop}A): 
\begin{equation} \label{eqg2}
    \dv{\loss}{w_{ij}} = \sum_{t} \dv{\loss}{\ovar^{t}_j} \cdv{ \ovar^{t}_j}{ \hvar^{t}_j}  \epsilon_{ij}^{t} = \sum_{t} \dv{\loss}{\ovar^{t}_j} e^{t}_{ij}
\end{equation} 

Even though $e^t_{ij}$ is causal and local, this equation as a whole is not, since the factor \smash{$\dvtext{\loss}{\ovar^{t}_j}$} still includes explicit recurrences. E-prop will simply ignore these recurrences to solve this problem (cf. Fig.~\ref{fig:eprop}B,  Section~\ref{chap:e-prop}).

\subsection{Re-expressing Explicit Recurrences of Order 1}

Now we re-express the explicit recurrences' gradient \smash{$\cdvtext{ \hvar_k^{t+1}}{ \ovar_j^{t}}$} analogously to the implicit recurrence in the previous section. First, we plug Eq.~\ref{eqdedz} (explicit recurrences) into Eq.~\ref{eqg2} (re-expressed implicit recurrence):
\begin{equation}\label{eqorder1}
\begin{split}\
    \dv{\loss}{w_{ij}} =& \sum_{t'} \left(\cdv{ \loss}{ \ovar^{t'}_j} + \sum_k \dv{\loss}{\hvar^{t'+1}_k} \cdv{ \hvar^{t'+1}_k}{ \ovar_j^{t'}}\right) e^{t'}_{ij} \\
    =& \sum_{t'} \cdv{ \loss}{ \ovar^{t'}_j} e^{t'}_{ij} + \sum_k \sum_{t'} \dv{\loss}{\hvar^{t'+1}_k} \cdv{ \hvar^{t'+1}_k}{ \ovar_j^{t'}} e^{t'}_{ij} 
\end{split}
\end{equation}

The first factor of this sum is already causal since it only requires the direct derivative and the implicit eligibility trace introduced in Eq.~\ref{eqiet}. Focusing on the second factor, this term represents the gradient until $\hvar^{t'+1}_k$, the jump to $\ovar^{t'}_j$ and the implicit eligibility trace $e^{t'}_{ij}$ stored there that represents the sum over all of the paths from there to $w_{ij}$. 

\subsubsection{Unrolling the recursion:} We can now unroll the recursion by plugging the equation of explicit recurrences Eq.~\ref{eqdedz} into the second term of Eq.~\ref{eqorder1}:
\begin{align}
    &\sum_{t'} \dv{\loss}{\hvar^{t'+1}_k} \cdv{ \hvar^{t'+1}_k}{ \ovar_j^{t'}} e^{t'}_{ij} = \sum_{t'} \left( \dv{\loss}{\ovar^{t'+1}_k} \cdv{ \ovar^{t'+1}_k}{ \hvar_k^{t'+1}} + \dv{\loss}{\hvar^{t'+2}_k}\cdv{ \hvar_k^{t'+2}}{ \hvar_k^{t'+1}} \right) \cdv{ \hvar^{t'+1}_k}{\ovar^{t'}_j}e^{t'}_{ij} \notag\\
    &= \sum_{t'} \left( \dv{\loss}{\ovar^{t'+1}_k} \cdv{ \ovar^{t'+1}_k}{ \hvar_k^{t'+1}} + \left( \dv{\loss}{\ovar^{t'+2}_k} \cdv{ \ovar^{t'+2}_k}{ \hvar_k^{t'+2}} + (\cdots) \cdv{ \hvar_k^{t'+3}}{ \hvar_k^{t'+2}} \right) \cdv{ \hvar_k^{t'+2}}{ \hvar_k^{t'+1}} \right) \cdv{ \hvar^{t'+1}_k}{\ovar^{t'}_j}e^{t'}_{ij} \notag \\
    &= \sum_{t'}\sum_{t \geq t'} \dv{\loss}{\ovar^{t+1}_k} \cdv{ \ovar^{t+1}_k}{ \hvar^{t+1}_k} \cdv{ \hvar^{t+1}_k}{ \hvar^{t}_k} \cdots \cdv{ \hvar^{t'+2}_k}{ \hvar^{t'+1}_k} \cdv{ \hvar^{t'+1}_k}{ \ovar^{t'}_j} e^{t'}_{ij}
\end{align}

\subsubsection{Flip time indices:} We flip the indices again to have a causal formula:
\begin{equation}\label{eqboim}
\begin{split}
    \sum_{t'} \dv{\loss}{\hvar^{t'+1}_k} \cdv{ \hvar^{t'+1}_k}{ \ovar_j^{t'}} e^{t'}_{ij}
    =& \sum_{t} \dv{\loss}{\ovar^{t+1}_k} \cdv{ \ovar^{t+1}_k}{ \hvar^{t+1}_k}  \sum_{t' \leq t} \cdv{ \hvar^{t+1}_k}{ \hvar^{t}_k} \cdots \cdv{ \hvar^{t'+2}_k}{ \hvar^{t'+1}_k} \cdv{ \hvar^{t'+1}_k}{ \ovar^{t'}_j} e^{t'}_{ij}
\end{split}
\end{equation}

\begin{figure}[t!]
\centering
\includegraphics[width=0.7\textwidth]{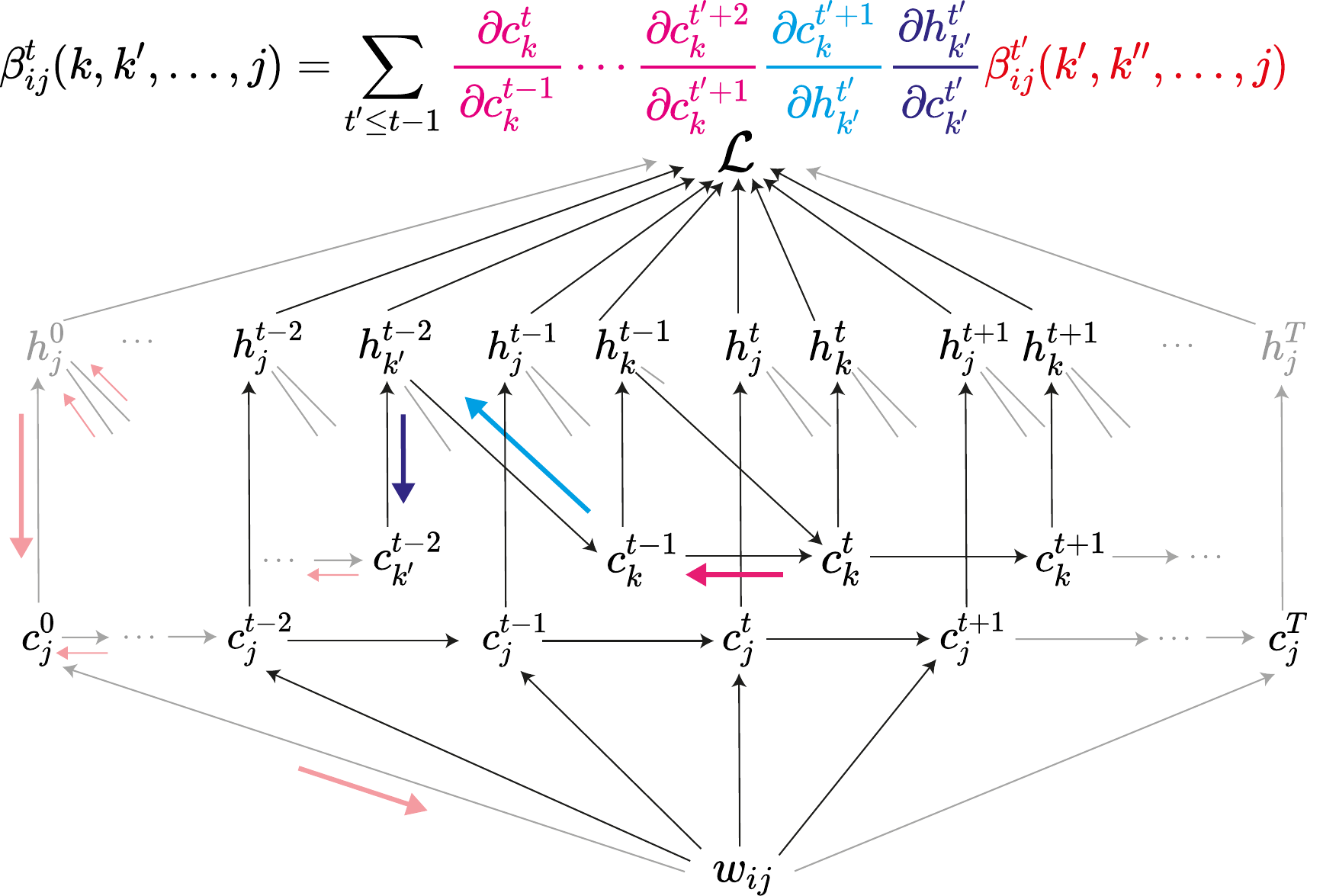} 
\caption{Computational graph for the explicit variable $\oimvar_{ij}^t(k,k',...,j)$ with $t'= t-1$.}
\label{fig:EET}
\end{figure}

\begin{definition*}[Explicit variable]
We define the \emph{explicit variable} $ \oimvar_{ij}^t(k,k',...,j)$ as:
\begin{equation}\label{eqeet}
     \oimvar_{ij}^t(k,k', ...,j) := \sum_{t' \leq t-1}  \cdv{\hvar^{t}_k}{ \hvar^{t-1}_k} \cdots \cdv{ \hvar^{t'+2}_k}{ \hvar^{t'+1}_k} \cdv{ \hvar^{t'+1}_k}{ \ovar^{t'}_{k'}} \cdv{ \ovar^{t'}_{k'}}{ \hvar^{t'}_{k'}} \oimvar_{ij}^{t'}(k', k'', ...,j)
\end{equation}
with $\oimvar_{ij}^t(j) = \epsilon^t_{ij}$.
\end{definition*}

\paragraph{Backwards interpretation:} The explicit variable represents the idea of starting at $\hvar^t_k$, moving an arbitrary number of steps through the implicit recurrence $\hvar^{t}_k \rightarrow h^{t-1}_k$ until at a certain $t'$ you jump to the output variable of another neuron $\ovar^{t'}_{k'}$, down to its hidden variable $\hvar^{t'}_{k'}$ and then start again, with a path, now starting at $\hvar^{t'}_{k'}$. In total, it considers all possible paths , with arbitrary length, spending an arbitrary number of steps in each of the neurons (through implicit recurrences) from $\hvar^t_k$ to $\hvar^{t'}_j$ through the neurons $k', k'', ...$ and then times the implicit variable $\epsilon^{t'}_{ij}$ (cf. Fig.~\ref{fig:EET}). 

\paragraph{Forwards interpretation:} The explicit variable accounts for the influence of the activity of neuron $j$ at any previous time step $\hvar^{t'}_j$ to neuron $k'$ at a future time step $\hvar^t_{k'}$ through the neurons $k', k'', ...$.

\paragraph{Incremental computation:} The recursive relation to this variable that allows it to be updated at each time step is:
\begin{equation}\label{eqincremeet}
      \oimvar_{ij}^t(k,k', ...,j) = \cdv{\hvar^{t}_k}{ \hvar^{t-1}_k} \oimvar_{ij}^{t-1}(k,k', ...,j) + \cdv{ \hvar^{t}_k}{ \ovar^{t-1}_{k'}} \cdv{ \ovar^{t-1}_{k'}}{ \hvar^{t-1}_{k'}} \oimvar_{ij}^{t-1}(k', ...,j)
\end{equation}

\begin{definition*}[Explicit eligibility trace]
Given the explicit variable $\oimvar_{ij}^t(k,k', ...,j)$, we define the \emph{explicit eligibility trace} $\oeltra_{ij}^t(k,k', ...,j)$ as:
\begin{equation}
    \oeltra_{ij}^t(k,k', ...,j) := \cdv{ \ovar^{t}_k}{ \hvar^{t}_k} \oimvar_{ij}^t(k,k', ...,j)
\end{equation} 
with $\oeltra_{ij}^t(j) = e^t_{ij}$

\end{definition*}

Since $\cdvtext{ \ovar^{t}_k}{ \hvar^{t}_k}$ is causal and local, and the explicit variable $\oimvar_{ij}^t(k,k', ...,j)$ is causal but only partially local (it requires message passing from the presynaptic neuron $k'$ to the postsynaptic neuron $k$), then the explicit eligibility trace $\oeltra_{ij}^t(k,k', ...,j)$ is also \textbf{causal} but only \textbf{partially local}.

\subsubsection{Final equation with re-expressed explicit recurrence of order 1:} Substituting the explicit variable Eq.~\ref{eqeet} in Eq.~\ref{eqboim} yields:
\begin{equation}\label{eqoimact}
\sum_k \sum_{t} \dv{\loss}{\hvar^{t+1}_k} \cdv{ \hvar^{t+1}_k}{ \ovar_j^{t}} e^{t'}_{ij} = \sum_k  \sum_{t} \dv{\loss}{\ovar^{t+1}_k} \oeltra_{ij}^{t+1}(k,j)
\end{equation}

And substituting this back to the original equation (cf. Eq.~\ref{eqorder1}):
\begin{equation}\label{eqorder1reex}
\begin{split}\
    \dv{\loss}{w_{ij}} =& \sum_{t} \cdv{ \loss}{ \ovar^{t}_j} e^{t}_{ij} + \sum_k \sum_{t} \dv{\loss}{\hvar^{t+1}_k} \cdv{ \hvar^{t+1}_k}{ \ovar_j^{t}} e^{t}_{ij}\\
    =& \sum_{t} \cdv{ \loss}{ \ovar^{t}_j} e^{t}_{ij} + \sum_k  \sum_{t} \dv{\loss}{\ovar^{t+1}_k} \oeltra_{ij}^{t+1}(k,j)
\end{split}
\end{equation}

Here it becomes clear how setting this second factor to $0$ is what gives us e-prop (cf. the right arrow in Fig.~\ref{fig:overview}), since we forcefully ignore the influence of a neuron on other neurons (and itself) through the explicit recurrences.

\subsection{Re-expressing Explicit Recurrences of Order $> 1$}

Now that we have seen how the explicit eligibility connects the activity of neuron $j$ with other neurons through explicit recurrences, we can use it to re-express higher-order explicit recurrences. 

\subsubsection{Unroll the recursion:} Starting from the equation with one order of explicit recurrence already re-expressed (cf. Eq.~\ref{eqorder1reex}), and alternatively using the definition of explicit recurrences (cf. Eq.~\ref{eqdedz}) and the action of the explicit eligibility trace (cf. Eq.~\ref{eqoimact}), we can repeat the previous steps for higher orders:
\begin{align}\label{eqorder2}
   \dv{\loss}{w_{ij}} =& \sum_{t} \cdv{ \loss}{ \ovar^{t}_j} e^{t}_{ij} + \sum_{k_1} \sum_{t} \left(  \cdv{ \loss}{ \ovar^{t+1}_{k_1}} + \sum_{k_2} \dv{\loss}{\hvar^{t+2}_{k_2}} \cdv{ \hvar^{t+2}_{k_2}}{ \ovar_{k_1}^{t+1}} \right) \oeltra_{ij}^{t+1}(k_1,j) \notag \\
   =& \sum_{t} \cdv{ \loss}{ \ovar^{t}_j} e^{t}_{ij} +  \sum_{t} \sum_{k_1} \cdv{ \loss}{ \ovar^{t+1}_{k_1}} \oeltra_{ij}^{t+1}(k_1,j) + \sum_{t} \sum_{k_1, k_2} \dv{\loss}{\hvar^{t+2}_{k_2}} \cdv{ \hvar^{t+2}_{k_2}}{ \ovar_{k_1}^{t+1}} \oeltra_{ij}^{t+1}(k_1,j) \notag \\
   =& \sum_{t} \cdv{ \loss}{ \ovar^{t}_j} e^{t}_{ij} +  \sum_{t} \sum_{k_1} \cdv{ \loss}{ \ovar^{t+1}_{k_1}} \oeltra_{ij}^{t+1}(k_1,j) + \sum_{t} \sum_{k_1, k_2} \dv{ \loss}{ \ovar^{t+2}_{k_2}} \oeltra_{ij}^{t+2}(k_2, k_1,j) \notag \\
   =& \sum_{t} \sum_{t' \geq t} \sum_{k_0=j, k_1,..,k_{t}} \cdv{ \loss}{ \ovar^{{t'}}_{k_{t}}} \oeltra_{ij}^{t'}(k_{t}, \cdots, k_1,k_0=j)
\end{align}

This gives us a high overview of separating the different levels of explicit recurrences which will lead to the definition of the m-order e-prop (Section~\ref{chap:e-prop}).

\subsubsection{Flip time indices:} As before we change the time indices and reorganise to allow for causality,
\begin{equation}\label{eqpchang}
\begin{split}
   \dv{\loss}{w_{ij}} =& \sum_{t} \sum_{k} \cdv{ \loss}{ \ovar^{{t}}_{k}} \cdv{ \ovar^{t}_{k}}{ \hvar^{t}_{k}} \sum_{t' \leq t} \sum_{k_0=j, k_1,..,k_{t'-1}}  \oimvar_{ij}^{t}(k, k_{t'-1}, \cdots, k_1,k_0 = j)
\end{split}
\end{equation}

\begin{definition*}[Recurrence variable]
%\paragraph{Definition:} 
We define the \emph{recurrence variable} $ \alpha^{t,r}_{ij}$ as: 
\begin{equation}\label{eqpet}
     \alpha^{t,r}_{ij} = \sum_{t' \leq t} \sum_{k_0=j, k_1,..,k_{t'-1}}  \oimvar_{ij}^{t}(r, k_{t'-1}, \cdots, k_1,k_0 = j)
\end{equation}
\end{definition*}

\paragraph{Backwards interpretation:} Starting at current time $t$ in neuron $r$, the recurrence variable represents all combinations of paths through any combination of neurons $k_{t'-1},..,k_1$ ending in neuron $j$.

\paragraph{Forwards interpretation:} The recurrence variable accounts for the influence of the activity of neuron $j$ at any previous time step to neuron $r$ at the current timestep $t$ through all possible paths through all neurons.

\paragraph{Incremental computation:} Once again, importantly, we have a recursive equation for computing the recurrence variable:  
\begin{equation}\label{eqincremret}
     \alpha^{t,r}_{ij} = \cdv{ \hvar^{t}_r}{ \hvar_{r}^{t-1}} \alpha^{{t-1},r}_{ij} + \sum_{k} \cdv{ \hvar^{t}_r}{ \ovar_{k}^{t-1}}
     \cdv{ \ovar^{t-1}_{k}}{\hvar_{k}^{t-1}} \alpha^{{t-1},k}_{ij} + \cdv{ \hvar^{t}_r}{w_{ij}}
\end{equation}
Observe $\forall r \neq j, \; \cdv{ \hvar^{t}_r}{w_{ij}} = 0$. 

\begin{definition*}[Recurrence eligibility trace]

Given the recurrence variable $\alpha^{t,r}_{ij}$, we define the \emph{recurrence eligibility trace} $a_{ij}^{t,r}$ as:
\begin{equation}
    a_{ij}^{t,r} := \cdv{ \ovar^t_{r}}{ \hvar_{r}^t} \alpha^{t,r}_{ij}
\end{equation}
\end{definition*}

Since \smash{$\cdvtext{ \ovar^t_{r}}{ \hvar_{r}^t}$} is causal and local, but the recurrence variable $\alpha^{t,r}_{ij}$ is causal but non-local, the recurrence eligibility trace $a^{t,r}_{ij}$ is also \textbf{causal} but \textbf{non-local}. It is non-local in an equivalent way as BPTT is not: each synapse $ij$ requires to store a variable representing how the activation in the past of any other neuron $r$ would affect its computation in the present, even if $r \neq i,j$ and through all possible paths of synapses. The recursive computation of $\alpha^{t,r}_{ij}$ requires of the summation of the recurrence variables of all the neurons requiring non-local communication.

\subsubsection{Final equation of RTRL:}

\begin{figure}[t!]
\centering
\includegraphics[width=0.65\textwidth]{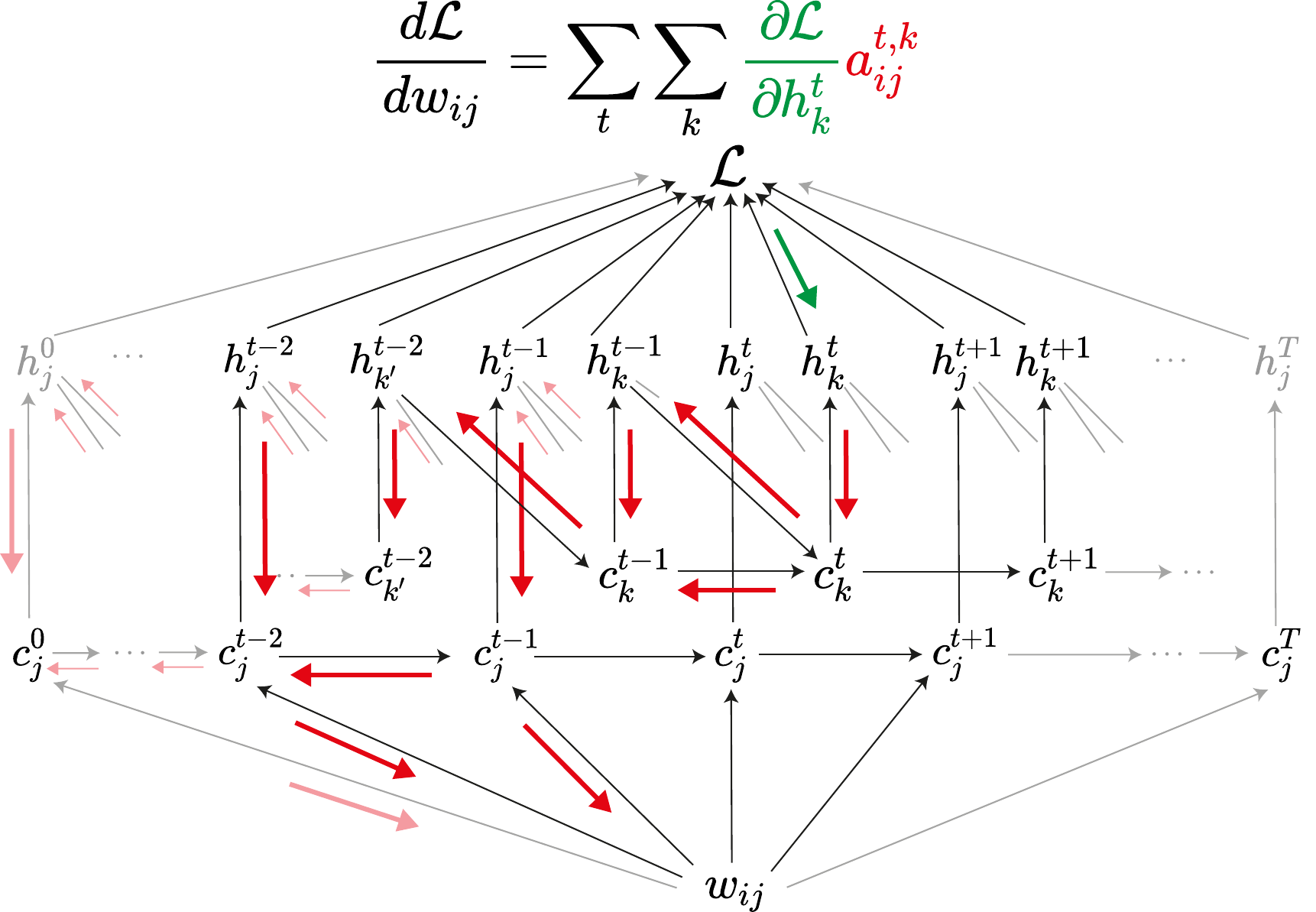} 
\caption{Computational graph for final computation of RTRL.} \label{fig:RTRL}
\end{figure}

Eq.~\ref{eqpchang} transforms into (by substituting Eq.~\ref{eqpet} into Eq.~\ref{eqpchang}) the final equation for RTRL (cf. Fig.~\ref{fig:RTRL}):
\begin{equation}\label{eqRTRL}
    \dv{\loss}{w_{ij}} = \sum_{t} \sum_{k} \cdv{ \loss}{ \ovar^{t}_{k}} \cdv{ \ovar^t_{k}}{ \hvar_{k}^t}  \alpha^{t,k}_{ij} = \sum_{t} \sum_{k} \cdv{ \loss}{ \ovar^{t}_{k}} a_{ij}^{t,k}
\end{equation}
We now have a causal but still non-local gradient computation algorithm.

\section{E-prop}
\label{chap:e-prop}

The e-prop algorithm approximates the gradient by not considering the explicit recurrences in RNNs. E-prop was originally formulated for RSNNs, since it is considered more biologically plausible than BPTT and RTRL due to its characteristics of being causal and local \cite{Bellec2020}. The approximation of the gradient that defines e-prop is: %
\begin{equation}\label{eqeprop}
    \dv{\loss}{w_{ij}} \approx \sum_{t} \cdv{\loss}{ \ovar^{t}_j} e^t_{ij}
\end{equation} 
Through the derivation of RTRL from BPTT, e-prop has arisen naturally in three different places. This allows us for equivalent interpretations of the approximation, each more detailed than the previous one. 

First, and as originally proposed \cite{Bellec2020}, we can understand e-prop from the equation that arises after re-expressing the implicit eligibility trace (cf. Eq.~\ref{eqg2}):
\begin{equation}
    \dv{\loss}{w_{ij}} = \sum_{t} \dv{\loss}{\ovar^{t}_j} e^t_{ij} \notag
\end{equation}

Here we approximate the non-causal and non-local total derivative by the causal and local partial derivative , i.e. \smash{$\dvtext{\loss}{\ovar^{t}_j} \approx \cdvtext{\loss}{ \ovar^{t}_j}$} (cf. Fig.~\ref{fig:eprop}).

Second, we can understand it from the equation after re-expressing the explicit recurrences of order 1 (cf. Eq.~\ref{eqorder1reex}):
\begin{equation}
    \dv{\loss}{w_{ij}} = \sum_{t} \cdv{ \loss}{ \ovar^{t}_j} e^{t}_{ij} + \sum_k  \sum_{t} \dv{\loss}{\ovar^{t+1}_k} \oeltra_{ij}^{t+1}(k,j) \notag
\end{equation}

Here we see explicitly what we are ignoring in the approximation, since ignoring this non-causal and non-local second term, i.e. \smash{$\sum_k  \sum_{t} \dv{\loss}{\ovar^{t+1}_k} \oeltra_{ij}^{t+1}(k,j)=0$}, defines e-prop. Ignoring these future dependencies to other neurons through explicit recurrences leads to a gradient computing algorithm that treats each neuron as producing an output only for the network's computation and not to communicate to other neurons. Therefore, synapses arriving at neurons that are not directly connected to the readout of the RNN, are not modified by e-prop (e-prop does not compute through additional feed-forward layers).

Finally, the most expressive of the interpretations comes from the equation that shows how to apply the re-expressing of the explicit recurrences of order 1, recursively, to re-express higher orders (cf. Eq.~\ref{eqorder2}):
\begin{align*}
   \dv{\loss}{w_{ij}} =& \sum_{t} \cdv{ \loss}{ \ovar^{t}_j} e^{t}_{ij} + \sum_k  \sum_{t} \dv{\loss}{\ovar^{t+1}_k} \oeltra_{ij}^{t+1}(k,j) \\ =& \sum_{t} \cdv{ \loss}{ \ovar^{t}_j} e^{t}_{ij} +  \sum_{t} \sum_{k_1} \cdv{ \loss}{ \ovar^{t+1}_{k_1}} \oeltra_{ij}^{t+1}(k_1,j) + \sum_{t} \sum_{k_1, k_2} \dv{ \loss}{ \ovar^{t+2}_{k_2}} \oeltra_{ij}^{t+2}(k_2, k_1,j) \\ 
   =& \cdots
\end{align*}

Here we define the \emph{m-order e-prop} as the approximation resulting from setting in the above equation $\sum_t \sum_{k_0 = j, k_1, ..., k_m} \dv{ \loss}{ \ovar^{t+m}_{k_m}} \oeltra_{ij}^{t+m}(k_m, \cdots, k_1,k_0=j)=0$. By increasing the order $m$ we  better approximate the gradient at the cost of needing the activities of other neurons $m$ time steps ahead to compute the current gradient of the loss.
Under this scope, standard e-prop \cite{Bellec2020} is just the 1-order e-prop (fully causal and local but the most inaccurate approximation). On the other extreme, the T-order e-prop (nothing is approximated or set to 0) corresponds to the full gradient computation, in a middle form between BPTT and RTRL (the exact computation of the gradient but completely non-causal and non-local).  Moreover, synapses arriving into neurons connected to the readout through up to $m-1$ synapses will be modified by the m-order e-prop (m-order e-prop computes through up to $m-1$ additional feed-forward layers).

\section{Conclusion}

In this paper, we formally explored how BPTT, RTRL, and e-prop relate to each other. We extended the general scheme for re-expressing recurrences as eligibility traces from \cite{Bellec2020} and applied it iteratively to go from BPTT to RTRL. In the process, we found intermediate expressions that allow for better intuition of these algorithms. Moreover, we showed how e-prop can be seen as an extreme case of a series of approximation algorithms, which we coin m-order e-prop. 

\bibliographystyle{splncs04}
\bibliography{2022-RecurrentLearningRules}

\clearpage

\newcommand{\rdv}[2]{\frac{d^r#1}{d^r#2}}
\newcommand{\rdvtext}[2]{\nicefrac{d^r#1}{d^r#2}}
\newcommand{\rimvar}[0]{\gamma}
\newcommand{\reltra}[0]{g}

\appendix

\chapter*{Appendix}
\vspace{-0.5cm}

\section{Online Computation vs Online Update}

In the main text, we have used $w_{ij}$ as a constant variable over the $T$ time steps. This is based on the usual usage of BPTT where we compute $\dvtext{\loss}{w_{ij}}$ after the $T$ time steps of execution and only then do the update $w^{T+1}_{ij} = w^0_{ij} + f(\dvtext{\loss}{w_{ij}})$. 

Looking at the RTRL equation Eq.~\ref{eqRTRL}:
\begin{equation}
    \dv{\loss}{w_{ij}} = \sum_{t} \sum_{k} \cdv{ \loss}{ \ovar^{t}_{k}} a_{ij}^{t,k}
\end{equation}
we observe that since the factors for a given $t$ are causal, we can compute the quantity $C^t = \sum_k \cdv{\loss}{\ovar^t_k}a^{t,k}_{ij}$ at each time step $t$ (while doing the forward pass) and add them all after $T$ time-steps to exactly compute $\dvtext{\loss}{w_{ij}} = \sum_t C^t$. Equivalently with e-prop and computing the value $C^t = \cdv{\loss}{h^t_j}e^t_{ij}$ to approximate $\dvtext{\loss}{w_{ij}} \approx \sum_t C^t$. This is why we consider these algorithms as \emph{online gradient computing algorithms}.

However, the exact equivalence between BPTT and these causal algorithms is only held if the weights are not updated with these $C^t$ values, since the derivations of these algorithms from BPTT depend on the weights being constant for the $T$ time steps. That is why in the computational graph we use $w_{ij}$ and not different $w^t_{ij} \; t\in{0,1,...,T}$ (cf. Fig.~\ref{fig:EandIandB}).

In general, we can define an \emph{online update algorithm} from each online gradient computation algorithm with the updating scheme $w^{t+1}_{ij} = w^t_{ij} + f(C^t)$ with $C^t$ being the causal values described before. In general, however, this is computationally different from BPTT, RTRL, and e-prop as described in this paper.

\section{E-prop with Read-Out Neurons}

When the RNN has a readout module, this has to be considered in e-prop in the computation of the gradient of the loss with respect to the output of the neuron, i.e. $\cdvtext{\loss}{\ovar^t_{ij}}$. Note that now, focusing on the read-out subnetwork, this derivative is not necessarily direct, so we will use a medium notation to refer to a derivative that has no paths into the main RNN but that is not completely direct. So $\cdvtext{\loss}{\ovar^t_{ij}}$ in the main text becomes $\rdvtext{\loss}{\ovar^t_{ij}}$ in this point of view.

We find ourselves in a subnetwork where we are computing $\rdvtext{\loss}{\ovar^t_{ij}}$. The BPTT equation (cf. Eq.~\ref{eqg1}) becomes in this case:

\begin{equation}\label{eqg1ron}
    \rdv{\loss}{\ovar^t_{j}} = \sum_{k} \rdv{\loss}{y_k^t} \cdv{y_k^t}{\ovar^t_{j}}
\end{equation}
where $y^t_k$ is the activity of the readout neuron $k$ at time $t$.

\subsection{No Recurrences} When the readout module is not recurrent (e.g. linear readout), we have that $\rdvtext{\loss}{y_k^t} = \cdvtext{\loss}{y_k^t}$ and therefore Eq.~\ref{eqg1ron} becomes the usual equation for Backpropagation for one layer:

\begin{equation}\label{eqron0}
    \rdv{\loss}{\ovar^t_{j}} = \sum_{k} \cdv{\loss}{y_k^t} \cdv{y_k^t}{\ovar^t_{j}}
\end{equation}

\subsubsection{Final e-prop equation with non-recurrent read-out:}
Substituting Eq.~\ref{eqron0} in the equation for e-prop Eq.~\ref{eqeprop}, we get the final equation for RNNs with a non-recurrent readout module:
\begin{equation}
    \dv{\loss}{w_{ij}} \approx \sum_{t} \sum_{k} \cdv{\loss}{y^t_k} \cdv{y_k^t}{\ovar^t_{j}} e^t_{ij}
\end{equation} 

\subsection{With Implicit Recurrences} When the readout module has implicit recurrences (e.g. non-spiking readout neurons), we have to use an implicit variable as in the main text (cf. Eq.~\ref{eqiet})

\begin{figure}[t!]
\includegraphics[width=\textwidth]{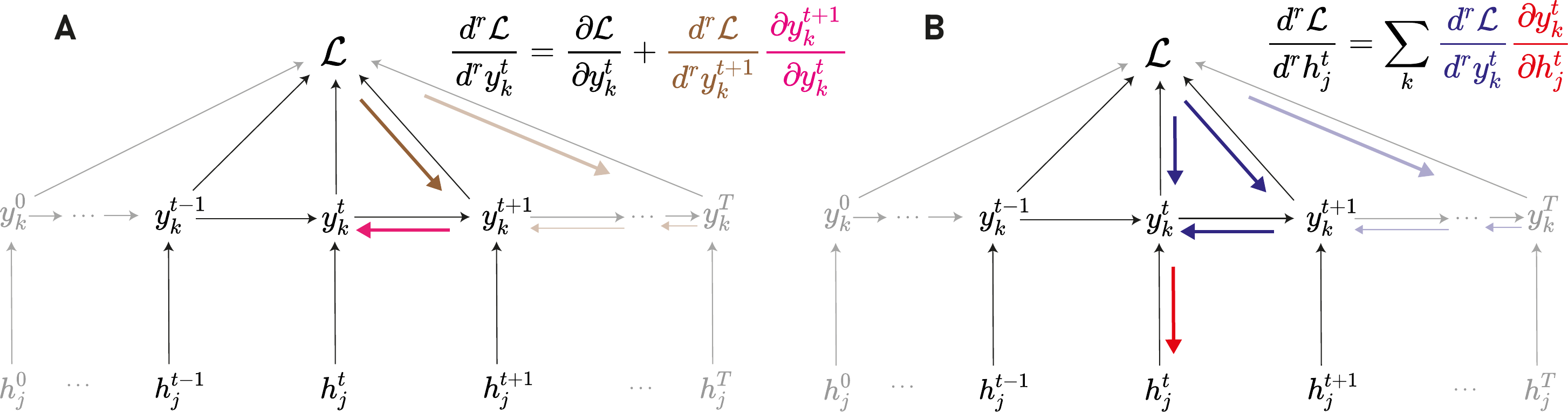} 
\caption{Computational graph for A) implicit recurrence gradients in read-out neurons B) computation of $\rdvtext{\loss}{ \ovar^t_j}$.}
\label{fig:Ron}
\end{figure}

Looking at the computational graph for $\rdvtext{\loss}{ \ovar^t_j}$ in this case (cf. Fig \ref{fig:Ron}B) we see we can do an equivalent trick as we did with the implicit recurrence in Section \ref{sect:ReexImplicitRec}. To do this, we first write a recursive definition of the gradient $\rdvtext{\loss}{y^t_k}$ (cf. Fig.~\ref{fig:Ron}A):

\begin{equation} \label{eqronimpl}
    \rdv{\loss}{y^t_k} = \cdv{\loss}{ y^t_k} + \rdv{\loss}{y^{t+1}_k} \cdv{ y^{t+1}_k}{ y^t_k}
\end{equation}

\subsubsection{Unrolling the recursion:} To unroll, we plug Eq.~\ref{eqronimpl} into Eq.~\ref{eqg1ron}:
\begin{equation} \label{eqronunroll}
\begin{split}
    \rdv{\loss}{ \ovar^{t'}_j} =& \sum_{k} \left( \cdv{\loss}{ y^{t'}_k} + \rdv{\loss}{y^{t'+1}_k}\cdv{ y_k^{t'+1}}{ y_k^{t'}} \right) \cdv{ y^{t'}_k}{ \ovar^{t'}_j}\\
    =&\sum_{k} \left( \cdv{\loss}{ y^{t'}_k} + \left( \cdv{\loss}{ y^{t'+1}_k} + (\dotsi) \cdv{ y_k^{t'+2}}{ y_k^{t'+1}} \right) \cdv{ y_k^{t'+1}}{ y_k^{t'}} \right) \cdv{ y^{t'}_k}{ \ovar^t_j}\\
    =&\sum_{k}\sum_{t \geq t'} \cdv{\loss}{ y^t_k} \cdv{ y^t_k}{ y^{t-1}_k} \dotsi \cdv{ y^{t'+1}_k}{ y^{t'}_k} \cdv{ y^{t'}_k}{ \ovar^{t'}_j}
\end{split}
\end{equation}

\subsubsection{Flip time indices:} Contrary to Section \ref{sect:ReexImplicitRec} with $w_{ij}$, in this case, we have $h^t_{ij}$ which changes over time, and so there is no summation over $t$ that would allow us for an exchange of time indices. However, we can introduce Eq.~\ref{eqronunroll} in the equation of symmetric e-prop Eq.~\ref{eqeprop}, which has a confluence of paths at $w_{ij}$ to do the time index flip:

\begin{equation}\label{eqreprop}
\begin{split}
    \dv{\loss}{w_{ij}} &\approx \sum_{t} \cdv{\loss}{ \ovar^{t}_j} e^t_{ij} = \sum_{t} \rdv{\loss}{ \ovar^{t}_j} e^t_{ij}\\
    &= \sum_k \sum_{t', t \geq t'} \cdv{\loss}{ y^t_k} \cdv{ y^t_k}{ y^{t-1}_k} \dotsi \cdv{ y^{t'+1}_k}{ y^{t'}_k} \cdv{ y^{t'}_k}{ \ovar^{t'}_j} e^{t'}_{ij}\\
    &= \sum_{k,t} \cdv{\loss}{ y^t_k} \sum_{t' \leq t} \cdv{ y^t_k}{ y^{t-1}_k} \dotsi \cdv{ y^{t'+1}_k}{ y^{t'}_k} \cdv{ y^{t'}_k}{ \ovar^{t'}_j} e^{t'}_{ij} 
\end{split}
\end{equation}

\begin{figure}[t!]
\centering
\includegraphics[width=0.75\linewidth]{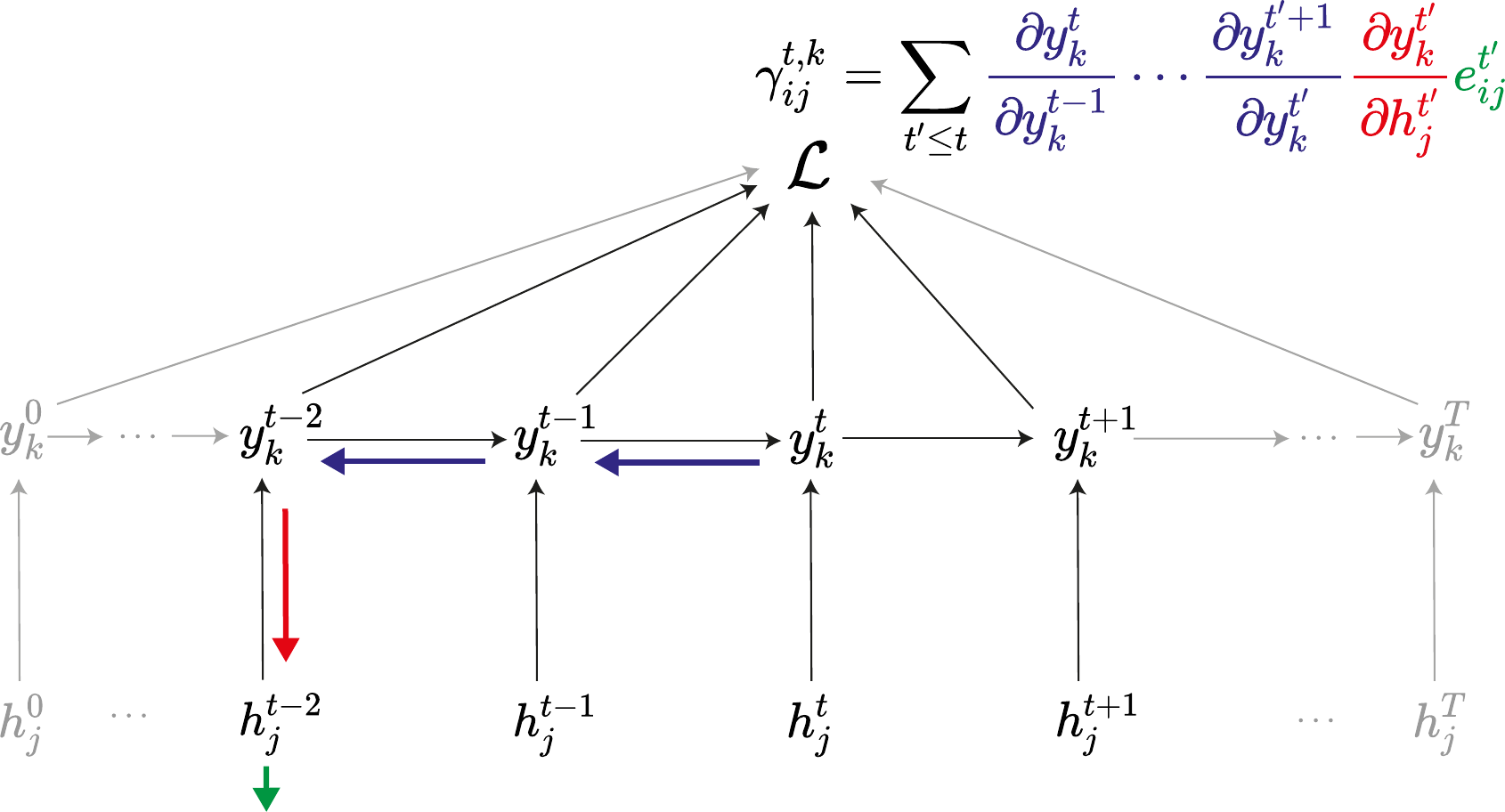} 
\caption{Computational graph for the read-out implicit variable $\rimvar^{t,k}_{ij}$ with $t'= t-2$.}
\label{fig:RIET}
\end{figure}

\begin{definition*}[Read-out implicit variable]
We define the \emph{read-out implicit variable} $\rimvar^t_{ij}$ as:
\begin{equation}\label{eqriet}
     \rimvar_{ij}^{t,k} := \sum_{t' \leq t} \cdv{ y^t_k}{ y^{t-1}_k} \cdots \cdv{ y^{t'+1}_k}{ y^{t'}_k} \cdv{ y^{t'}_k}{h^{t'}_j} e^{t'}_{ij}
\end{equation}

\end{definition*}

\paragraph{Backwards interpretation:} Starting at $y^t_k$, the read-out implicit variable represents the sum over all the paths going backwards through the implicit recurrence until $y^{t'}_k$ and the connection to all other paths in the RNN (without explicit recurrences since we are using e-prop) stored there in $e^{t'}_{ij}$ (cf. Fig.~\ref{fig:RIET}).

\paragraph{Forwards interpretation:} The read-out implicit variable represents how the read-out neuron's value $y^t_k$ has been affected by the states of the RNN through time, i.e. taking into account also how the values of the read-out neuron at previous time steps have affected the value at the current time step through the implicit recurrence. 

\paragraph{Incremental computation:} 
The recursive relation:
\begin{equation}\label{eqincremriet}
     \rimvar_{ij}^{t,k} = \cdv{ y^t_k}{ y^{t-1}_k}  \rimvar^{t-1,k}_{ij} + \cdv{ y^t_k}{ \ovar^t_j} e^t_{ij}
\end{equation}

\begin{definition*}[Read-out implicit eligibility trace]
Given the read-out implicit variable $\rimvar^{t,k}_{ij}$, we define the \emph{read-out implicit eligibility trace} $\reltra^{t,k}_{ij}$ as:
\begin{equation}
    \reltra^{t,k}_{ij} := \cdv{\loss}{ y^t_k} \rimvar^{t,k}_{ij}
\end{equation} 
\end{definition*}

Since $\cdvtext{\loss}{ y^t_k}$ is causal, and so is the read-out implicit variable $\rimvar^t_{ij}$ (can be computed at each time step), then the read-out implicit eligibility trace $\reltra^t_{ij}$ is also \textbf{causal}. However, it is \textbf{semi-local} since it requires information both from the connection $jk$ and $ij$.

\subsubsection{Final e-prop equation with read-out re-expressed implicit recurrence:}
With all of this combined, substituting Eq.~\ref{eqriet} in Eq.~\ref{eqreprop} we get the following: 
\begin{equation} \label{eqrong2}
    \dv{\loss}{w_{ij}} \approx \sum_{k,t} \cdv{\loss}{ y^t_k} \rimvar_{ij}^{t,k}  = \sum_{k,t} \reltra^{t,k}_{ij}
\end{equation}

\subsection{LSNNs}

For completeness, we repeat here the reformulation of read-out neurons with implicit recurrence in the original e-prop paper \cite{Bellec2019}. Particularly, they use as read-out non-spiking neurons with update equation:

\begin{equation}\label{eqron}
    y^t_k = \kappa y^{t-1}_k + \sum_j w^{out}_{jk}\ovar^t_j + b_k^{out}
\end{equation}
where $\kappa$ is the read-out neurons' time constant, $w^{out}_{jk}$ is the $jk$ connection's strength and $b_k^{out}$ is a bias.
 
 From the definition of read-out implicit variable Eq.~\ref{eqriet} and using $\cdvtext{ y^t_k}{ y^{t-1}_k} = \kappa$ and $\cdvtext{ y^{t}_k}{ \ovar^{t}_j} = w^{out}_{jk}$ (cf. Eq.~\ref{eqron}) we conclude:

\begin{equation}
     \rimvar_{ij}^{t,k} = \sum_{t' \leq t} \kappa^{t'-t} w^{out}_{jk} e^{t'}_{ij}
\end{equation}
that can be rewritten as:

\begin{equation}
     \rimvar_{ij}^{t,k} = w^{out}_{jk} \mathcal{F}_{\kappa}(e^t_{ij})
\end{equation}
where $\mathcal{F}_{\alpha}(x^t) = \alpha\mathcal{F}_{\alpha}(x^{t-1}) + x^t$ is a \emph{low-pass filter operator} for a time series $x$.

Finally, if we plug this in the final e-prop equation for read-outs with implicit recurrence Eq.~\ref{eqrong2}:
\begin{equation} 
    \dv{\loss}{w_{ij}} \approx \sum_{k,t} \cdv{\loss}{ y^t_k} \rimvar_{ij}^{t,k}  = \sum_{k,t} \cdv{\loss}{ y^t_k} w^{out}_{jk} \mathcal{F}_{\kappa}(e^t_{ij}) = \sum_{t} L^t_j \mathcal{F}_{\kappa}(e^t_{ij})
\end{equation} 
where $L^t_j = \sum_k w^{out}_{jk} \cdv{\loss}{ y^t_k}$ is called a \emph{learning signal}.

\section{Complexity}

E-prop solves the non-locality problem of RTRL (except for the learning signal) while being causal. Also, it is less computational and memory expensive than RTRL. Given $n$ neurons simulated $T$ time steps with $p$ synapses, we see the complexities of the different algorithms in Table \ref{table:com} \cite{Bellec2020,Zenke2020}:

\begin{table}[b!]
\centering
\caption{Computational and memory complexity of BPTT, RTRL and e-prop.} 
\label{table:com}
\begin{tabularx}{\linewidth}{@{\extracolsep{30pt}}CCC}
\toprule   
{} & \multicolumn{2}{c}{Complexity}\\
 \cmidrule{2-3} 
 Algorithm & Computation & Memory \\ 
\midrule
BPTT & $O(pT)$ & $O(nT)$  \\ 
RTRL & $O(p^2T)$ & $O(np)$ \\ 
e-prop & $O(pT)$ & $O(p)$ \\ 
\bottomrule
\end{tabularx}
\end{table}

Notice that in general in dense architectures $p = n^2$.
RTRL has a computational complexity of $O(p^2T)$ in total ($O(p^2)$ per time step). This complexity is prohibitive in comparison to BPTT which belongs to $O(pT)$ in total ($O(p)$ per time step, even though since it is not causal it cannot be implemented online and so there is no per time step computation). E-prop has a computational complexity of $O(pT)$ in total (and indeed now $O(p)$ per time step), so is as efficient as BPTT.

With regards to memory complexity, BPTT is $O(nT)$ (the variables' values of each neuron at each time step, since it is not local), usually much less than the memory requirements of RTRL which is $O(np)$ (the recurrent eligibility variable of each neuron for each synapse $\alpha^{t,r}_{ij}$). E-prop has a memory complexity of $O(p)$ (the implicit eligibility variable of each synapse $\epsilon^t_{ij}$). This is not as good as BPTT but way better than RTRL.

The main drawback of e-prop is that is an approximation of the gradient and, therefore, in theory, it should take more iterations to converge to an optimal value. In practical examples the increase in the number of iterations is not prohibitive \cite{Bellec2019,Bellec2020}.

\section{Proofs of incremental computations}

\subsection{Proof of the Incremental Computation of the Implicit Variable}

Incremental computation of the implicit variable Eq.~\ref{eqincremiet}:
\begin{equation*}
     \epsilon_{ij}^t = \cdv{ \hvar^t_j}{ \hvar^{t-1}_j}  \epsilon^{t-1}_{ij} + \cdv{ \hvar^t_j}{ w_{ij}}
\end{equation*}

\subsubsection{Proof:}
Using the definition of the implicit variable Eq.~\ref{eqiet}
\begin{equation*}
     \epsilon_{ij}^t := \sum_{t' \leq t} \cdv{ \hvar^t_j}{ \hvar^{t-1}_j} \cdots \cdv{ \hvar^{t'+1}_j}{ \hvar^{t'}_j} \cdv{ \hvar^{t'}_j}{ w_{ij}}
\end{equation*}
we get
\begin{equation*}
\begin{split}
     \epsilon_{ij}^t &= \cdv{ \hvar^t_j}{ \hvar^{t-1}_j}  \sum_{t' \leq t-1} \cdv{ \hvar^{t-1}_j}{ \hvar^{t-2}_j} \cdots \cdv{ \hvar^{t'+1}_j}{ \hvar^{t'}_j} \cdv{ \hvar^{t'}_j}{ w_{ij}} + \cdv{ \hvar^t_j}{ w_{ij}} \\
     &= \sum_{t' \leq t-1} \cdv{ \hvar^t_j}{ \hvar^{t-1}_j}   \cdv{ \hvar^{t-1}_j}{ \hvar^{t-2}_j} \cdots \cdv{ \hvar^{t'+1}_j}{ \hvar^{t'}_j} \cdv{ \hvar^{t'}_j}{ w_{ij}} + \cdv{ \hvar^t_j}{ w_{ij}} \\
     &= \sum_{t' \leq t} \cdv{ \hvar^t_j}{ \hvar^{t-1}_j} \cdots \cdv{ \hvar^{t'+1}_j}{ \hvar^{t'}_j} \cdv{ \hvar^{t'}_j}{ w_{ij}}
\end{split}
\end{equation*}
\qed\\

Equivalent proof for the incremental computation of the read-out implicit variable Eq.~\ref{eqincremriet}.

\subsection{Proof of the Incremental Computation of the Explicit Variable}

Incremental computation of the explicit variable Eq.~\ref{eqincremeet}:
\begin{equation*}
      \oimvar_{ij}^t(k,k', ...,j) = \cdv{\hvar^{t}_k}{ \hvar^{t-1}_k} \oimvar_{ij}^{t-1}(k,k', ...,j) + \cdv{ \hvar^{t}_k}{ \ovar^{t-1}_{k'}} \cdv{ \ovar^{t-1}_{k'}}{ \hvar^{t-1}_{k'}} \oimvar_{ij}^{t-1}(k', ...,j)
\end{equation*}

\subsubsection{Proof:}
Using the definition of the explicit variable Eq.~\ref{eqeet} 
\begin{equation*}
     \oimvar_{ij}^t(k,k', ...,j) := \sum_{t' \leq t-1}  \cdv{\hvar^{t}_k}{ \hvar^{t-1}_k} \cdots \cdv{ \hvar^{t'+2}_k}{ \hvar^{t'+1}_k} \cdv{ \hvar^{t'+1}_k}{ \ovar^{t'}_{k'}} \cdv{ \ovar^{t'}_{k'}}{ \hvar^{t'}_{k'}} \oimvar_{ij}^{t'}(k', k'', ...,j)
\end{equation*}
we get
\begin{equation*}
\begin{split}
     \oimvar_{ij}^t(k,k', ...,j) &= \cdv{\hvar^{t}_k}{ \hvar^{t-1}_k} \sum_{t' \leq t-2}  \cdv{\hvar^{t-1}_k}{ \hvar^{t-2}_k} \cdots \cdv{ \hvar^{t'+2}_k}{ \hvar^{t'+1}_k} \cdv{ \hvar^{t'+1}_k}{ \ovar^{t'}_{k'}} \cdv{ \ovar^{t'}_{k'}}{ \hvar^{t'}_{k'}} \oimvar_{ij}^{t'}(k', k'', ...,j)\\
     &+ \cdv{ \hvar^{t}_k}{ \ovar^{t-1}_{k'}} \cdv{ \ovar^{t-1}_{k'}}{ \hvar^{t-1}_{k'}} \oimvar_{ij}^{t-1}(k', ...,j)\\
     &= \sum_{t' \leq t-2} \cdv{\hvar^{t}_k}{ \hvar^{t-1}_k}   \cdv{\hvar^{t-1}_k}{ \hvar^{t-2}_k} \cdots \cdv{ \hvar^{t'+2}_k}{ \hvar^{t'+1}_k} \cdv{ \hvar^{t'+1}_k}{ \ovar^{t'}_{k'}} \cdv{ \ovar^{t'}_{k'}}{ \hvar^{t'}_{k'}} \oimvar_{ij}^{t'}(k', k'', ...,j)\\
     &+ \cdv{ \hvar^{t}_k}{ \ovar^{t-1}_{k'}} \cdv{ \ovar^{t-1}_{k'}}{ \hvar^{t-1}_{k'}} \oimvar_{ij}^{t-1}(k', ...,j)\\
     &= \sum_{t' \leq t-1}  \cdv{\hvar^{t}_k}{ \hvar^{t-1}_k} \cdots \cdv{ \hvar^{t'+2}_k}{ \hvar^{t'+1}_k} \cdv{ \hvar^{t'+1}_k}{ \ovar^{t'}_{k'}} \cdv{ \ovar^{t'}_{k'}}{ \hvar^{t'}_{k'}} \oimvar_{ij}^{t'}(k', k'', ...,j)
\end{split}
\end{equation*}
\qed

\subsection{Proof of the Incremental Computation of the Recurrence Variable}

Incremental computation of the recurrence variable Eq.~\ref{eqincremret}:
\begin{equation*}
     \alpha^{t,r}_{ij} = \cdv{ \hvar^{t}_r}{ \hvar_{r}^{t-1}} \alpha^{{t-1},r}_{ij} + \sum_{k} \cdv{ \hvar^{t}_r}{ \ovar_{k}^{t-1}}
     \cdv{ \ovar^{t-1}_{k}}{\hvar_{k}^{t-1}} \alpha^{{t-1},k}_{ij}  + \cdv{ \hvar^{t}_r}{w_{ij}}
\end{equation*}

\subsubsection{Proof:}
Using the definition of the explicit variable Eq.~\ref{eqeet} 
\begin{equation*}
      \oimvar_{ij}^t(k,k', ...,j) = \cdv{\hvar^{t}_k}{ \hvar^{t-1}_k} \oimvar_{ij}^{t-1}(k,k', ...,j) + \cdv{ \hvar^{t}_k}{ \ovar^{t-1}_{k'}} \cdv{ \ovar^{t-1}_{k'}}{ \hvar^{t-1}_{k'}} \oimvar_{ij}^{t-1}(k', ...,j)
\end{equation*}
and the definition of the recurrence variable Eq.~\ref{eqpet} 
\begin{equation*}
     \alpha^{t,r}_{ij} = \sum_{t' \leq t} \sum_{k_0=j, k_1,..,k_{t'-1}}  \oimvar_{ij}^{t}(r, k_{t'-1}, \cdots, k_1,k_0 = j)
\end{equation*}
we get
\begin{equation*}
    \begin{split}
        \alpha^{t,r}_{ij} =& \cdv{ \hvar^{t}_r}{ \hvar_{r}^{t-1}} \sum_{t' \leq t-1} \sum_{k_0=j, k_1,..,k_{t'-1}}  \oimvar_{ij}^{t-1}(r, k_{t'-1}, \cdots, k_1,k_0 = j)\\
        +& \sum_{k} \cdv{ \hvar^{t}_r}{ \ovar_{k}^{t-1}}
     \cdv{ \ovar^{t-1}_{k}}{\hvar_{k}^{t-1}} \sum_{t' \leq t-1} \sum_{k_0=j, k_1,..,k_{t'-1}}  \oimvar_{ij}^{t-1}(k, k_{t'-1}, \cdots, k_1,k_0 = j)\\ 
     +& \cdv{ \hvar^{t}_r}{w_{ij}} \\
     =& \cdv{ \hvar^{t}_r}{ \hvar_{r}^{t-1}} \sum_{0 < t' \leq t-1} \sum_{k_0=j, k_1,..,k_{t'-1}}  \oimvar_{ij}^{t-1}(r, k_{t'-1}, \cdots, k_1,k_0 = j) \\
        +& \sum_{k} \cdv{ \hvar^{t}_r}{ \ovar_{k}^{t-1}}
     \cdv{ \ovar^{t-1}_{k}}{\hvar_{k}^{t-1}} \sum_{t' < t-1} \sum_{k_0=j, k_1,..,k_{t'-1}}  \oimvar_{ij}^{t-1}(k, k_{t'-1}, \cdots, k_1,k_0 = j)\\
     +& \left( \cdv{ \hvar^{t}_r}{ \hvar_{r}^{t-1}} \oimvar_{ij}^{t-1}(r) + \cdv{ \hvar^{t}_r}{w_{ij}} \right) \\
     +& \sum_{k} \cdv{ \hvar^{t}_r}{ \ovar_{k}^{t-1}}
     \cdv{ \ovar^{t-1}_{k}}{\hvar_{k}^{t-1}} \sum_{k_0=j, k_1,..,k_{t-2}}  \oimvar_{ij}^{t-1}(k, k_{t-2}, \cdots, k_1,k_0 = j)\\
     =& \sum_{t' \leq t-1} \sum_{k_0=j, k_1,..,k_{t'-1}}  \oimvar_{ij}^{t}(r, k_{t'-1}, \cdots, k_1,k_0 = j)\\
     +& \sum_{k} \cdv{ \hvar^{t}_r}{ \ovar_{k}^{t-1}}
     \cdv{ \ovar^{t-1}_{k}}{\hvar_{k}^{t-1}} \sum_{k_0=j, k_1,..,k_{t-2}}  \oimvar_{ij}^{t-1}(k, k_{t-2}, \cdots, k_1,k_0 = j)
    \end{split} 
\end{equation*}

Now, since $\oimvar_{ij}^{t-1}(r,k, k_{t-2}, \cdots, k_1,k_0 = j) = 0 \quad \forall r,k,k_{t-2}, \cdots, k_1$ (since it considers $t$ explicit jumps in $t-1$ steps), we can add it to the equation:

\begin{equation*}
    \begin{split}
        \alpha^{t,r}_{ij} =& \sum_{t' \leq t-1} \sum_{k_0=j, k_1,..,k_{t'-1}}  \oimvar_{ij}^{t}(r, k_{t'-1}, \cdots, k_1,k_0 = j)\\
     +& \cdv{ \hvar^{t}_r}{ \hvar_{r}^{t-1}} \sum_{k_0=j, k_1,..,k_{t-2}}  \oimvar_{ij}^{t-1}(r,k, k_{t-2}, \cdots, k_1,k_0 = j)\\
     +& \sum_{k} \cdv{ \hvar^{t}_r}{ \ovar_{k}^{t-1}}
     \cdv{ \ovar^{t-1}_{k}}{\hvar_{k}^{t-1}} \sum_{k_0=j, k_1,..,k_{t-2}}  \oimvar_{ij}^{t-1}(k, k_{t-2}, \cdots, k_1,k_0 = j)\\
     =& \sum_{t' \leq t-1} \sum_{k_0=j, k_1,..,k_{t'-1}}  \oimvar_{ij}^{t}(r, k_{t'-1}, \cdots, k_1,k_0 = j)\\
     +& \sum_{k_0=j, k_1,..,k_{t-1}}  \oimvar_{ij}^{t}(r,k_{t-1}, k_{t-2}, \cdots, k_1,k_0 = j)\\
     =& \sum_{t' \leq t} \sum_{k_0=j, k_1,..,k_{t'-1}}  \oimvar_{ij}^{t}(r, k_{t'-1}, \cdots, k_1,k_0 = j)
    \end{split} 
\end{equation*}

\qed

\end{document}